\documentclass[letterpaper]{article} 
\usepackage[preprint]{aaai2027}

\usepackage[hyphens]{url}  
\usepackage{graphicx} 
\usepackage{natbib}  
\usepackage{caption} 
\usepackage{algorithm}
\usepackage{algorithmic}
\usepackage{amsmath}
\usepackage{amsthm}
\usepackage{amssymb}
\newtheorem{proposition}{Proposition}
\newtheorem{corollary}{Corollary}
\newtheorem{definition}{Definition}
\theoremstyle{remark}
\newtheorem{remark}{Remark}
\newtheorem{assumption}{Assumption}

\usepackage{newfloat}
\usepackage{listings}
\DeclareCaptionStyle{ruled}{labelfont=normalfont,labelsep=colon,strut=off} 
\floatstyle{ruled}
\newfloat{listing}{tb}{lst}{}
\floatname{listing}{Listing}

\usepackage{booktabs}

\title{CalibratedRubric: Task-Adaptive Rubric Banks for Open-Ended LLM Evaluation}
\author{
    Mengting Chen\textsuperscript{\rm 1},
    Yanshu Sun\textsuperscript{\rm 1},
    Wanting Liang\textsuperscript{\rm 1},
    Beidi Luan\textsuperscript{\rm 2},
    Rui Sun\textsuperscript{\rm 2},
    Dezhi Chen\textsuperscript{\rm 1},
    Jing Li\textsuperscript{\rm 2},
    Zuo Bai\textsuperscript{\rm 2,\rm 1}\corresponding
}
\affiliations{
    \textsuperscript{\rm 1}FinStep\\
    \textsuperscript{\rm 2}StepFun\\
    baizuo@stepfun.com, baizuo@finstep.cn
}

\begin{document}

\maketitle

\begin{abstract}
Reliable evaluation of open-ended LLM outputs requires fine-grained rubrics, yet expert curation is costly and difficult to scale. Existing automated pipelines rely on strict judge unanimity and binary variance filters, which cannot distinguish measurable rubrics from informative ones. We introduce \emph{CalibratedRubric}, a task-adaptive framework that combines type-specific scoring, Bayesian rubric-measurability filtering, and item response theory (IRT)-based bank assembly. CalibratedRubric estimates each rubric's measurability with a Beta--Bernoulli agreement posterior and uses a submodular information-coverage objective to construct compact rubric banks over the observed capability range. Across financial, healthcare, general, and legal benchmarks, measurability filtering improves human-gold agreement on JudgmentBench from $\kappa=0.604$ to $0.743$. IRT-based greedy selection improves cross-fitted rank fidelity over random selection across all six evaluated response blocks and requires only 49 rather than 131 rubrics to reach the target correlation on FinResearchBench decision-support tasks. Task-label perturbations further reduce system separation, confirming the practical relevance of task-adaptive scoring. These results support CalibratedRubric as an efficient, uncertainty-aware approach to open-ended LLM evaluation, with calibration gains depending on sufficient judge redundancy.
\end{abstract}

\section{Introduction}
\label{sec:intro}

LLM-based autonomous agents increasingly adopt decoupled, service-oriented designs with dedicated memory, retrieval-augmented generation (RAG), and tool-use layers, enabling long-form outputs over extended interaction horizons~\cite{wang2024,xi2023}.
This shift makes conventional automatic metrics inadequate: lexical-overlap measures such as BLEU and ROUGE and embedding-based measures such as BERTScore were designed for comparatively short, reference-bounded outputs and correlate weakly with expert judgment on open-ended deliverables such as synthesized financial research reports~\cite{papineni2001,lin,zhang2020}.
The LLM-as-a-Judge paradigm offers a scalable alternative, but
holistic scores conflate distinct quality dimensions and remain
susceptible to length, position, and self-preference biases
~\cite{li2023,zheng2023,liu2023geval,wang2024fairevaluator}. Analytic evaluators therefore decompose quality into skill-level,
rubric-conditioned, multidimensionally calibrated, collaboratively
aligned, or atomic-checklist judgments
~\cite{ye2024,kim2024,hashemi2024,chiang2026,cook2024}.

Two structural bottlenecks limit current analytic evaluators.
First, reliable rubric authoring still depends heavily on scarce domain expertise.
For example, constructing a financial benchmark to professional examination standards requires experts to draft, refine, and validate criteria, making annotation difficult to scale~\cite{liu2026}.
Second, automated methods can synthesize instance-specific
criteria~\cite{wang2026,jia2026,li2026}, but evaluation items vary
substantially in informativeness~\cite{rodriguez2021}, and reliable
selection remains largely heuristic. Prior work uses IRT to build NLP evaluation scales, compare test sets,
and shrink benchmarks~\cite{lalor2016,vania2021,polo2024}; we instead
embed IRT inside rubric construction and use the item information
function to select criteria over the relevant capability range
~\cite{birnbaum1968,lord1980,baker2001}.
The consensus-derived pipeline we extend retains only rubrics on which all judges agree and applies a binary variance filter~\cite{luan}.
This favors reproducibility but treats judges as equally reliable and criteria as equally informative. A binary filter removes only completely constant criteria and cannot quantify how much the remaining ones distinguish systems.

We propose \emph{CalibratedRubric}, replacing fixed deterministic pipelines with task-adaptive, probabilistically calibrated evaluation.
A task-typing front end maps each query to a cognitive task type---evidence reasoning, decision support, high-risk constraint, or divergent creation---and assigns an appropriate rubric form and aggregation rule.
A Bayesian calibration module models judge-specific leniency and reliability as latent parameters, replacing strict unanimity with robust posterior consensus anchored by a small human-labeled subset.
Finally, we embed item response theory (IRT) inside rubric construction: candidate rubrics are treated as items, and the item information function (IIF) selects criteria that maximize information over the relevant capability range~\cite{birnbaum1968,lord1980,baker2001}.

Our contributions are fourfold:
(1)~We introduce a task-adaptive taxonomy that dynamically selects rubric forms and scoring rules based on cognitive demands.
(2)~We deploy IRT and the IIF as an information-maximizing rubric-selection mechanism, producing system capability estimates with calibrated confidence intervals.
(3)~We develop a Bayesian judge-calibration procedure that models bias and reliability, establishing robust posterior consensus.
(4)~Across general and specialized domains, we demonstrate that the framework preserves human-expert alignment while dramatically compressing evaluation costs and improving the resolution of closely matched systems.

\section{Problem Setup and Theoretical Analysis}
\label{sec:setup}

\subsection{Problem Formulation}
\label{subsec:notation}

Let $\mathcal{Q}=\{q_n\}_{n=1}^{N}$ be the queries and
$\mathcal{M}=\{m_i\}_{i=1}^{M}$ the evaluated systems, with output
$r_{in}=m_i(q_n)$.  For each query, a generator produces a
query-specific pool $\mathcal{C}_n$ of natural-language criteria;
$\mathcal{C}=\bigsqcup_n\mathcal{C}_n$ and $n(j)$ denotes the query of
criterion $j$.  Judge $k\in\{1,\ldots,K\}$ assigns
$y^{(k)}_{ij}\in\mathcal{S}_j$ to output $r_{i,n(j)}$, and
$\bar y_{ij}$ is the aggregated label.

We use $z_j\in\{0,1\}$ for \emph{reproducible judgeability}: whether
competent graders can apply criterion $j$ consistently.  This is
necessary but not sufficient for substantive expert endorsement, a
gap evaluated on the human-labelled subset.  Without additional
elicitation, judge $k$ supplies the validity vote
$v^{(k)}_j=\mathbf{1}[y^{(k)}_{ij}=\bar
y^{(-k)}_{ij}\ \forall i]$, where $\bar y^{(-k)}$ is the
leave-one-judge-out majority.

Given a retained set $\mathcal{G}\subseteq\mathcal{C}$ and
non-negative weights $w_j$, system $i$ receives
\begin{equation}
S(m_i)=\sum_{n=1}^{N}\omega_n\,
\frac{\sum_{j\in\mathcal{G}\cap\mathcal{C}_n} w_j\, g_j(\bar{y}_{ij})}
     {\sum_{j\in\mathcal{G}\cap\mathcal{C}_n} w_j},
\qquad \textstyle\sum_n \omega_n = 1,
\label{eq:score}
\end{equation}
where $g_j$ maps the response scale to $[0,1]$.  The within-query
normalization prevents queries with larger candidate pools from
dominating the score.  Gold-set construction must balance
\emph{validity}, \emph{discriminative information}, and
\emph{parsimony}, since judging costs
$O(KM|\mathcal{G}|)$.  We therefore solve
\begin{equation}
\mathcal{G}^{\star}=\!\!\arg\max_{\mathcal{G}\subseteq\mathcal{C},\,|\mathcal{G}|\le k}\!\!
\mathcal{I}(\mathcal{G})
\;\;\text{s.t.}\;\;
\min_{j\in\mathcal{G}} \Pr\!\left(z_j=1\mid Y\right)\ge\tau_c ,
\label{eq:problem}
\end{equation}
where $\mathcal{I}$ is test information and $\tau_c$ is a
validity threshold. Both quantities are latent and are estimated from
judge labels and a small human anchor set.

\begin{definition}[Consensus-derived baseline]
\label{def:baseline}
The baseline \cite{luan} uses uniform weights and retains item $j$ iff
\begin{align}
\delta^{\mathrm{cons}}_j&=\mathbf{1}\!\left[\,y^{(k)}_{ij}=y^{(k')}_{ij}\;
\forall i\in\mathcal{M},\;\forall k,k'\in\mathcal{K}\,\right],
\label{eq:cons}\\
\delta^{\mathrm{disc}}_j&=\mathbf{1}\!\left[\,\exists\, i,i' :\;
\bar{y}_{ij}=1,\;\bar{y}_{i'j}=0 \,\right],
\label{eq:disc}
\end{align}
are both one.
\end{definition}

The two hard rules correspond to the constraint and objective of
\eqref{eq:problem}, respectively.  We next show why both are crude
approximations; derivations and diagnostics appear in
Appendix~\ref{app:theory}.

\subsection{Unanimity Attrition}
\label{subsec:consensus}

\begin{assumption}[Judge error model]
\label{as:judge}
Conditional on the expert label of an (item, deliverable) pair, judges label independently, and judge $k$ returns the expert label with probability $1-\varepsilon_k$.
\end{assumption}

\begin{proposition}[Exponential attrition]
\label{prop:attrition}
Under Assumption~\ref{as:judge} with homogeneous error rate $\varepsilon$, for any item $j$ the unanimity rule \eqref{eq:cons} satisfies
\begin{equation}
\Pr\!\left(\delta^{\mathrm{cons}}_j=1\right)=
\Big[(1-\varepsilon)^{K}+\varepsilon^{K}\Big]^{M}=:\rho(\varepsilon,K)^{M}.
\label{eq:attrition}
\end{equation}
\end{proposition}

The factor $\rho^M$ suppresses ambiguous items, but it arbitrarily removes informative and uninformative criteria at the same rate, worsening as the leaderboard grows. 

\begin{proposition}[Reliability-aware consensus]
\label{prop:bayes}
Let judge $k$ have sensitivity $s_k^+$ and specificity $s_k^-$ \cite{dawid1979}.  The validity posterior is the weighted vote
\begin{equation}
\begin{split}
\log\frac{\Pr(z_j{=}1\mid Y)}{\Pr(z_j{=}0\mid Y)}
=\;&\log\frac{\pi_0}{1-\pi_0}
+\sum_{k}\Big[ v^{(k)}_{j}\log\frac{s^{+}_k}{1-s^{-}_k}\\
&+\big(1-v^{(k)}_{j}\big)\log\frac{1-s^{+}_k}{s^{-}_k}\Big],
\end{split}
\label{eq:logodds}
\end{equation}
Thresholding \eqref{eq:logodds} is Bayes-optimal for $0$--$1$ validity loss.
\end{proposition}

Thus posterior consensus decouples retention from literal unanimity, allowing reliable judges to carry appropriate weight. Appendix~\ref{app:assumptions} details the diagnostic protocol and identifiability conditions.

\subsection{Variance Filtering}
\label{subsec:iif}

Under the two-parameter logistic (2PL) model \cite{birnbaum1968,lord1980}, the probability that system $i$ satisfies binary criterion $j$ is
\begin{equation}
P_{ij}=\Pr\!\left(y_{ij}{=}1\mid \theta_i\right)=\sigma\!\big(a_j(\theta_i-b_j)\big),
\label{eq:2pl}
\end{equation}
where $\sigma(u)=(1+e^{-u})^{-1}$, and $a_j$ and $b_j$ are the discrimination and difficulty of criterion $j$. The item information function and the test information of a set $\mathcal{G}$ are
\begin{equation}
I_j(\theta)=a_j^2 P_j(\theta)\big(1{-}P_j(\theta)\big),
\quad I_{\mathcal{G}}=\!\!\sum_{j\in\mathcal{G}}\! I_j,
\label{eq:iif}
\end{equation}
and the standard error of the ability estimate is $\mathrm{SE}(\hat{\theta})=I_{\mathcal{G}}(\theta)^{-1/2}$.
We instantiate the objective in \eqref{eq:problem} as the information delivered where the evaluated systems actually lie, $\mathcal{I}(\mathcal{G})=\int I_{\mathcal{G}}(\theta)\varphi(\theta)\,d\theta$, with $\varphi$ the estimated capability distribution of $\mathcal{M}$.

\begin{remark}[Degenerate information filter]
\label{prop:degenerate}
Let $\hat{p}_j=\frac{1}{M}\sum_i \bar{y}_{ij}$. Then
\begin{equation}
\delta^{\mathrm{disc}}_j
=\mathbf{1}\!\left[\widehat{\mathrm{Var}}\big(\bar{y}_{\cdot j}\big)>0\right]
=\mathbf{1}\big[\hat{I}_j>0\big],
\label{eq:degenerate}
\end{equation}
where $\hat I_j=a^2\hat p_j(1-\hat p_j)$ assumes common
discrimination and ignores system ability.  Hence the baseline is the
zero-threshold, ability-blind special case of information selection.
\end{remark}

This heuristic cannot rank surviving criteria, distinguish
ability-consistent from aberrant response patterns, or target
difficulties to the capability range.  The IIF supplies all three and
supports a fixed budget.

\begin{proposition}[Strict refinement]
\label{prop:refine}
Let the item parameters be estimated by marginal maximum likelihood with a proper prior on $b$, let items with constant observed responses be assigned $\hat{I}_j\equiv 0$, and let
$\mathcal{G}_{\mathrm{IIF}}(k)=\arg\max_{|\mathcal{G}|\le k,\, \hat{a}_j> 0}\int I_{\mathcal{G}}(\theta)\varphi(\theta)d\theta$.
For any $k\le|\{j:\widehat{\mathrm{Var}}(\bar{y}_{\cdot j})>0\}|$,
$\mathcal{G}_{\mathrm{IIF}}(k)\subseteq\{j:\delta^{\mathrm{disc}}_j=1\}$,
with strict inclusion whenever $k$ is smaller than the number of surviving items or some surviving item has $\hat{a}_j\le 0$.
\end{proposition}

Proposition~\ref{prop:refine} states the relationship we claim precisely: information-based selection never retains a criterion the baseline heuristic rejects, and prunes further among those it accepts. This predicts that the resulting ranking agrees with the baseline ranking while requiring strictly fewer criteria to reach the same separation, the hypothesis we test experimentally. The screen $\hat{a}_j>0$ is a sign convention, not a tuned threshold: it removes items whose fitted response function decreases in ability, which contribute rank noise rather than information. We use it at $0$ throughout and never tune it, so the budget $B$ and the validity threshold $\tau_c$ remain the only quantities set per dataset.

On a purely binary block the feasible sets can coincide; any gain then
comes from ranking items and stopping early, not from admitting new
items.  We therefore test budgeted separation rather than claim a
binary validity gain.  The proof and tightness conditions are in
Appendix~\ref{app:iif-proof}.

\subsection{Task-Adaptive Measurement}
\label{subsec:scale}

Binary criteria are adequate for verifying whether an evidential step is present; they are a poor instrument for grading the quality of a recommendation, and they cannot express the asymmetric cost of a compliance violation. We therefore introduce a cognitive task type $t\in\mathcal{T}$, where $\mathcal{T}$ consists of four categories---\emph{evidence-reasoning}, \emph{decision-support}, \emph{safety-critical} and \emph{divergent-creative}---together with a classifier $\Pr(t\mid q)$ and a map from task type to a triple of response scale, measurement model and selection objective, summarized in Table~\ref{tab:tasktypes}.

\begin{table}[t]
\centering
\small
\setlength{\tabcolsep}{3pt}
\caption{Task-adaptive scoring rules used in the main experiments.}
\label{tab:tasktypes}
\begin{tabular}{@{}lll@{}}
\toprule
\textbf{Task type} & \textbf{Scoring} & \textbf{Task emphasis} \\
\midrule
Evidence & binary & pass $=1$, fail $=0$ \\
Decision & weighted binary & $5\times$ actionability \\
High-risk & risk-adjusted & $-0.3/-0.5$ violations \\
Creative & weighted binary & $3\times$ creativity \\
\bottomrule
\end{tabular}
\end{table}

The mapping changes both the response scale and the objective. Ordered responses can be more efficient when quality is genuinely graded, whereas safety-critical tasks call for expected violation loss rather than symmetric information. Table~\ref{tab:tasktypes} defines the comprehensive taxonomy of this task-adaptive design. In our main empirical evaluations, we rigorously demonstrate the maximum-information selection mechanism, proving its high robustness, scalability, and compression efficiency across diverse domains. Further theoretical motivation for the expanded branches appears in Appendix~\ref{app:task-scale}.

\paragraph{Assumptions and Scope.}
\label{subsec:assumptions}
The analysis assumes a dominant latent dimension, local item
independence, conditionally independent judges, and identifiable
ability estimates.  We stabilize small-$M$ estimation with priors,
compare 1PL and 2PL explicitly, and report bootstrap uncertainty.
Residual-spectrum and judge-correlation checks are reported
experimentally; the full assumptions and diagnostics are collected in
Appendix~\ref{app:assumptions}.

\section{CalibratedRubric}
\label{sec:method}

\begin{figure*}[t]
\centering
\includegraphics[width=\textwidth]{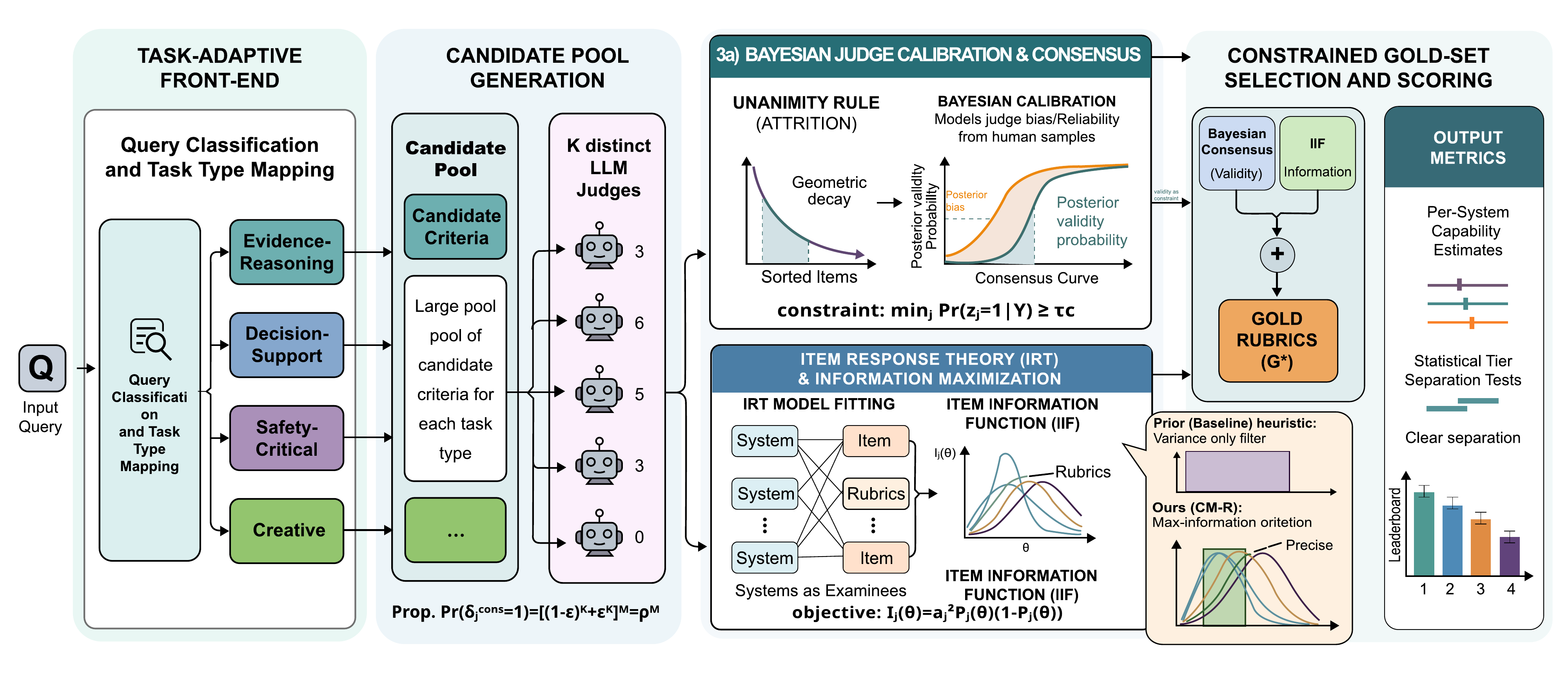}
\caption{CalibratedRubric. Task typing determines the scoring rule; Bayesian rubric calibration supplies the measurability constraint; IRT supplies item information; and budgeted assembly returns a compact rubric bank with uncertainty-aware system scores.}
\label{fig:framework}
\end{figure*}

Section~\ref{sec:setup} formulates rubric-bank construction as the constrained program in \eqref{eq:problem}. CalibratedRubric estimates its two ingredients from judge outputs: posterior rubric measurability and discriminative information. The former replaces literal unanimity with a probabilistic retention rule, while the latter supports budgeted rubric selection and uncertainty-aware capability estimation. Figure~\ref{fig:framework} summarizes the pipeline.

We take the candidate pools $\mathcal{C}$ as given and optimize which rubrics survive, how many are needed, and how they are weighted. The method is generator-agnostic but cannot recover dimensions absent from the candidate pool. Its stages type each task, align rubric scales, estimate measurability and information, and assemble a budgeted rubric bank.

\subsection{Task Typing}
\label{subsec:stage1}

A prompted classifier estimates $\Pr(t\mid q_n)$ over the four types in \S\ref{subsec:scale}. The query-level prediction determines the shared response scale and measurement family for $\mathcal{C}_n$ through Table~\ref{tab:tasktypes}. If $\max_t\Pr(t\mid q_n)<\tau_t$, the method falls back to binary maximum-information scoring; ordered responses are examined only in the GRM sensitivity analysis.

\subsection{Candidate Rubrics}
\label{subsec:stage2}

Each candidate is a (predicate, scale) pair. Scale mismatches are resolved by the deterministic map of \S\ref{subsec:scale}: graded items are binarized at their midpoint when required, and binary items become two-level graded items otherwise. Selection occurs only in \eqref{eq:problem}; $B$ and $\tau_c$ are the dataset-level controls.

\subsection{Bayesian Rubric Calibration}
\label{subsec:judge}
Existing LLM-judge calibration methods primarily correct
response-level selection or positional biases
~\cite{wang2024fairevaluator,li2025calibraeval}.
Ours instead estimates rubric-level measurability from
repeated inter-judge agreement.

For rubric $j$, let $n^{(j)}_{\mathrm{agree}}$ denote the number of evaluated instances on which all available judges agree and let $n^{(j)}_{\mathrm{total}}$ denote the total number of evaluated instances. We place a Beta prior on its latent measurability probability,
\begin{equation}
q_j\sim\operatorname{Beta}(\alpha,\beta),
\end{equation}
which yields the posterior
\begin{equation}
q_j\mid Y
\sim
\operatorname{Beta}\!\left(
\alpha+n^{(j)}_{\mathrm{agree}},
\beta+n^{(j)}_{\mathrm{total}}-n^{(j)}_{\mathrm{agree}}
\right).
\end{equation}
The posterior mean
\begin{equation}
\hat q_j
=
\mathbb{E}[q_j\mid Y]
=
\frac{\alpha+n^{(j)}_{\mathrm{agree}}}
{\alpha+\beta+n^{(j)}_{\mathrm{total}}}
\label{eq:measurability}
\end{equation}
provides a soft alternative to literal unanimity. We use the uniform prior $\alpha=\beta=1$ and retain rubrics satisfying $\hat q_j\geq\tau_c$; their instance-level labels are then aggregated by majority voting.

This estimator requires no human labels and is the calibration procedure evaluated in Section~\ref{sec:experiment}. Proposition~\ref{prop:bayes} describes a judge-specific extension for settings with sufficient expert anchors; it is not instantiated in the present experiments. With only two judges, unanimity coincides with pairwise agreement and therefore provides no independent rubric-level calibration signal.

\subsection{Item Calibration}
\label{subsec:irt}

\paragraph{Model and estimation.} Binary items use logistic IRT and ordered items use the graded response model \cite{samejima1969}. For binary data we compare 1PL and 2PL fits. With $\Delta\ell$ and $\Delta p$ denoting the richer model's gain and additional parameters, define
\begin{equation}
\kappa \;=\; \Delta\ell/\Delta p ,
\label{eq:kappa}
\end{equation}
AIC selects 2PL when $\kappa>1$ and BIC when $\kappa>\tfrac12\log N_{\mathrm{obs}}$; we report both because item-specific slopes are weakly identified on small leaderboards. Parameters are estimated by marginal MAP on 41 nodes over $[-4,4]$, followed by EAP ability estimation \cite{bock1981}. We impose $\theta\sim\mathcal N(0,1)$, $a_j>0$, and $\log a_j\sim\mathcal N(0,0.5^2)$; the last constraint prevents perfectly separating items from acquiring divergent discrimination and dominating the selector. Constant-response items receive zero information.

\paragraph{Information.} With the fitted parameters, the per-item scalar objective is the information delivered where the systems actually lie,
\begin{equation}
\nu_j=\int I_j(\theta)\,\varphi(\theta)\,d\theta ,
\qquad
\mathcal{I}(\mathcal{G})=\sum_{j\in\mathcal{G}}\nu_j ,
\label{eq:nu}
\end{equation}
where $\varphi$ is the fitted ability density. For ordered items, $I_j$ is the graded-response information; this branch is used only as a sensitivity analysis. We report residual-spectrum concentration and bootstrap-calibrated item fit. Confirmatory MIRT, Yen's $Q_3$~\cite{yen1984}, and dependent-item
refitting are left to future work.

\subsection{Rubric Assembly}
\label{subsec:select}

The measurability gate defines
\begin{equation}
\mathcal{F}
=
\{j:\hat q_j\geq\tau_c,\ I_j\not\equiv0\},
\label{eq:feasible-set}
\end{equation}
where $\hat q_j$ is the posterior measurability score in \eqref{eq:measurability}.
Plain IIF, analogous to classical information-based test
assembly~\cite{vanderlinden2000}, statically ranks feasible items by
\begin{equation}
\bar I_j=\sum_{g=1}^{G}\pi_g I_j(\theta_g)
\label{eq:plain-iif}
\end{equation}
and takes the top $B$, where $\pi_g$ are normalized standard-normal quadrature weights. It does not account for information already supplied by selected items.

Our final assembler instead optimizes the concave information-coverage
utility
\begin{equation}
U(S)=\sum_{g=1}^{G}\pi_g\log\!\left(1+I_S(\theta_g)\right),
\qquad
I_S(\theta_g)=\sum_{j\in S}I_j(\theta_g).
\label{eq:submod-utility}
\end{equation}
Starting from $S=\varnothing$, \textsc{IIF-Greedy} repeatedly selects
\begin{align}
j^\star&=\arg\max_{j\in\mathcal{F}\setminus S}\Delta U(j\mid S),
\label{eq:submod-greedy}\\
\Delta U(j\mid S)
&=\sum_{g=1}^{G}\pi_g
\log\!\left(1+\frac{I_j(\theta_g)}
{1+I_S(\theta_g)}\right),
\nonumber
\end{align}
and updates $S\leftarrow S\cup\{j^\star\}$ until $|S|=B$.

\begin{proposition}[Submodular assembly guarantee]
\label{prop:greedy}
The utility in \eqref{eq:submod-utility} is normalized, monotone
non-decreasing, and submodular.  Under the cardinality constraint
$|S|\le B$, the greedy rule \eqref{eq:submod-greedy} satisfies
\begin{equation}
U(S_{\mathrm{greedy}})
\ge (1-1/e)\max_{S\subseteq\mathcal{F},\,|S|\le B}U(S).
\end{equation}
\end{proposition}

Because test information is modular and $\log(1+x)$ is increasing and concave, the utility is normalized, monotone, and submodular; the standard greedy guarantee follows
~\cite{nemhauser1978analysis}. The denominator in \eqref{eq:submod-greedy} discounts already-covered ability regions, inducing diversity without an explicit penalty. The guarantee is conditional on local independence.

\paragraph{Small-$M$ interpretation.} When 1PL is selected, $\nu_j$ is maximized by difficulties near the centre of the fitted ability distribution. Thus the defensible gains on small leaderboards are difficulty targeting, budget awareness, and coverage, rather than precise recovery of item-specific discrimination.

\paragraph{Weights and scores.} The retained items are scored through \eqref{eq:score} with $g_j$ the scale normalizer of \S\ref{subsec:notation} and weights
\begin{equation}
w_j \;\propto\; \nu_j ,
\label{eq:weights}
\end{equation}
so informative rubrics contribute more, while posterior measurability remains solely the feasibility constraint defining $\mathcal{F}$. This separation prevents measurability from being counted twice and isolates the contribution of information weighting.

\paragraph{Uncertainty and tiers.} Query-stratified item bootstrap yields percentile intervals for $\hat\theta_i$. Adjacent systems whose ability difference is not significant are collapsed into a tier, avoiding unsupported fine-grained rankings.

\begin{algorithm}[t]
\caption{CalibratedRubric}
\label{alg:cmr}
\begin{algorithmic}[1]
\REQUIRE responses $Y$, candidate pools $\{\mathcal C_n\}$,
budget $B$, threshold $\tau_c$
\ENSURE rubric bank $\mathcal G^\star$, weights $w$,
capability tiers
\STATE type each query and align rubric scales
\STATE compute posterior measurability scores $\hat q_j$
\STATE fit regularized IRT and evaluate $I_j(\theta_g)$
\STATE $\mathcal F\leftarrow
\{j:\hat q_j\geq\tau_c,\ I_j\not\equiv0\}$
\STATE $S\leftarrow\varnothing$
\FOR{$b=1,\ldots,B$}
  \STATE $j^\star\leftarrow
  \arg\max_{j\in\mathcal F\setminus S}\Delta U(j\mid S)$
  \STATE $S\leftarrow S\cup\{j^\star\}$
\ENDFOR
\STATE $\mathcal G^\star\leftarrow S$;
set $w_j\propto\nu_j$
\STATE bootstrap rubrics and report capability tiers
\end{algorithmic}
\end{algorithm}

\paragraph{Cost and Amortization.}
Initial calibration costs $O(KMJ)$ judge calls, IRT fitting costs $O(\mathrm{iter}\,MJG)$, and greedy assembly costs $O(BJG)$ for $G=41$ cached nodes. A new system is then evaluated on $|\mathcal{G}^\star|$ items rather than the full $J$, converting repeated full-pool evaluation into a one-off calibration cost. We therefore report the smallest bank that preserves the full-pool ranking, not only fixed-budget accuracy.

\section{Experiment}
\label{sec:experiment}

This section evaluates the task-adaptive front end of
\emph{CalibratedRubric} on four datasets:
\textsc{FinResearchBench-v2}~\cite{luan},
\textsc{HealthBench}~\cite{arora2025healthbench},
\textsc{HelloBench}~\cite{que2024hellobench}, and
\textsc{JudgmentBench}~\cite{yang2026judgmentbench}.

\subsection{Task-Adaptive Scoring}
\textbf{Experimental Setup.}
This section evaluates the task-adaptive front end of ConsensusMultiRubric on four datasets: \textsc{FinResearchBench-v2} (hereafter \textsc{FinResearchBench}) as the main ranking domain, \textsc{HealthBench} and \textsc{HelloBench} as transfer settings, and \textsc{JudgmentBench} as a legal-domain validation set with human annotations and output-quality strata. Although the full collection contains 5,781 tasks and 65,648 rubrics, the current E1 snapshot uses 534 tasks and 7,959 rubrics with available responses (Appendix~\ref{app:e1-tables}).

Each task is mapped to one of four cognitive types---evidence reasoning, decision support, high-risk constraint, or creative divergent---which determines the scoring rule. The current pipeline removes invalid judge outputs, applies discrimination filtering, and yields 5,077 rubric entries after valid-output pruning and 2,813 final gold rubrics. Adaptive scoring uses task-specific rules, whereas the baseline applies a uniform binary scorer.

\textbf{Results and Analysis.}
The current main experiment yields 2,813 gold rubrics and ranks 21 systems, with Gemini first, \textit{Research Report} last, and a top-bottom gap of 94.26 percentage points. The strongest evidence for task typing comes from causal ablation: when 30\% of task labels are corrupted, the top-bottom gap drops from 94.86 to 83.95, a degradation of 10.91 points.

\begin{table}[t]
\centering
\small
\setlength{\tabcolsep}{4pt}
\begin{tabular}{@{}r r r r@{}}
\toprule
\textbf{Flip} & \textbf{Avg Score} & \textbf{Gap (pp)} & \textbf{Deg.} \\
\midrule
0.0 & 0.4413 & 94.86 & baseline \\
0.1 & 0.4419 & 94.86 & 0.00 \\
0.2 & 0.4284 & 86.97 & -7.89 \\
0.3 & 0.3922 & 83.95 & -10.91 \\
\bottomrule
\end{tabular}
\caption{Causal ablation under perturbed task-type assignments.}
\label{tab:e1-ablation}
\end{table}

Beyond causal usefulness, adaptive scoring remains ranking-stable and yields moderate discrimination gains, especially in robust distributional metrics such as IQR and standard deviation. On FinResearch, both adaptive and binary baselines achieve the same external validity against the human reference ranking ($\rho=0.8833$; Appendix~\ref{app:e1-tables}). On \textsc{JudgmentBench}, rubric-based scores recover a monotonic quality trend across tiers, even though within-task fine-grained ordering remains weak.

Overall, E1 shows that task-adaptive scoring is operationally stable, causally meaningful, and moderately beneficial for discrimination.

\subsection{Bayesian Judge Calibration}

L1 estimates a posterior measurability score for each rubric using the Beta--Bernoulli posterior mean
\begin{equation}
\mathbb{E}[q_r]
=
\frac{\alpha+n_{\mathrm{agree}}}
{\alpha+\beta+n_{\mathrm{total}}},
\end{equation}
where $n_{\mathrm{agree}}$ counts instances on which all available judges agree, $n_{\mathrm{total}}$ is the number of evaluated instances, and $\alpha=\beta=1$. Rubrics satisfying $\mathbb{E}[q_r]\geq\theta$ are retained and evaluated by majority voting. Importantly, L1 selection uses only inter-judge agreement and requires neither human labels nor a held-out gold judge.

We sweep $\theta\in\{0.50,0.60,0.65,0.70,0.80\}$ against unfiltered majority voting. Agreement increases monotonically as the threshold becomes stricter, at the expected cost of lower coverage. From $\theta=0.50$ to $0.80$, $\kappa$ rises from $0.8885$ to $0.9708$ for FinResearch decision support, from $0.8994$ to $0.9732$ for evidence reasoning, and from $0.661$ to $0.743$ against human gold on \textsc{JudgmentBench}. Table~\ref{tab:type_stratified_results} reports the strictest setting.

\begin{table}[t]
\centering
\small
\setlength{\tabcolsep}{2.5pt}
\caption{Agreement after L1 filtering at $\theta=0.80$. ``Kept'' reports retained rubrics where available.}
\label{tab:type_stratified_results}
\resizebox{\columnwidth}{!}{%
\begin{tabular}{@{}lllcccc@{}}
\toprule
\textbf{Task} & \textbf{Data} & \textbf{Gold}
& \textbf{Kept} & \textbf{Base} & \textbf{L1} & \textbf{$r$} \\
\midrule
Decision & FinResearch & LLM
& 286/624 & 0.8513 & 0.9708 & 0.589 \\
Evidence & FinResearch & LLM
& 81/195 & 0.8829 & 0.9732 & 0.558 \\
High-risk & Judgment & Human
& --- & 0.6040 & 0.7430 & 0.127 \\
\bottomrule
\end{tabular}
}
\end{table}

The FinResearch subsets further show that L1 preferentially removes unreliable rubrics rather than merely reducing coverage. For decision support, the bottom and top measurability quartiles have mean $\kappa$ values of $0.498$ and $0.983$; the corresponding values for evidence reasoning are $0.556$ and $0.953$. Consistently, posterior measurability correlates with agreement in both subsets ($r=0.589$ and $0.558$). The weaker correlation on the predominantly high-risk \textsc{JudgmentBench} ($r=0.127$) suggests that measurability is less predictive under human--LLM disagreement, although filtering still improves human-gold agreement from $0.604$ to $0.743$.

L1 requires at least three judges to provide an informative rubric-level signal: with two judges, unanimity is identical to pairwise agreement and cannot independently calibrate reliability. Accordingly, \textsc{HealthBench} and \textsc{HelloBench} show no additional gain under their two-judge configurations. Evaluation coverage also remains limited for evidence-reasoning and creative-divergent tasks. Moreover, LLM judges have higher positive-label rates (55.6--62.9\%) than the JudgmentBench human gold (47.1\%), indicating a systematic judge--human mismatch that measurability filtering does not fully eliminate.

\subsection{Submodular IRT Test Assembly and Capability Calibration}
\label{subsec:irt-experiment}

\textbf{Experimental Setup.}
We evaluate six response blocks covering all four task types, with
254--2,142 rubrics per block. The FinResearch blocks contain 15
systems judged by three LLMs; three transfer blocks contain six systems
and four judges; JudgmentBench contains 1,314 quality-stratified
outputs evaluated by a five-member human--LLM panel. Binary labels are
majority aggregated, while ordered labels are used only for GRM
sensitivity analysis.

Each block uses the regularized 2PL assembler in
\S\ref{subsec:irt}, with 41 nodes on $[-4,4]$ and
$\log a_j\sim\mathcal N(0,0.5^2)$. We compare \textsc{IIF-Greedy}
against plain IIF, unrestricted random selection, and random sampling
within the hard-filtered set. AIC, BIC, held-out likelihood, a
100-replicate item-fit bootstrap, and 300 item bootstraps diagnose
model fit and ability uncertainty.

Rank fidelity is Spearman's $\rho$ against the full usable block,
summarized as AUC over 12 logarithmic budgets. For sample-out
evaluation, rubrics are randomly half-split: selection and refitting use half A, while the reference ability vector is estimated from half B. We use 20 splits and three random draws per split; the target minimum-bank correlation is .95, or .9429 for six systems.

\textbf{Results and Analysis.}
AIC and BIC select 1PL in five of six blocks; JudgmentBench alone
splits (AIC: 2PL; BIC: 1PL), while held-out likelihood selects 3PL only for the creative block. Thus the regularized 2PL assembler is best interpreted as an operational mechanism for difficulty targeting and coverage over the observed ability range. Bootstrap item misfit ranges
from 0.23\% to 18.91\%, with decision support the clearest warning case; confirmatory MIRT and $Q_3$ remain future diagnostics.

On binary blocks, non-constancy and the hard discrimination filter
define the same feasible set, so the substantive question is
compression rather than set overlap. The hard filter is especially
brittle on heterogeneous outputs, retaining zero high-risk and four
creative items.

\begin{table*}[t]
\centering
\small
\setlength{\tabcolsep}{4pt}
\begin{tabular}{@{}llccccc@{}}
\toprule
\textbf{Task type} & \textbf{Response pool} &
\textbf{Greedy} & \textbf{Hard} & \textbf{Random} &
\textbf{$\Delta$Hard (95\% CI)} & \textbf{$\Delta$Random (95\% CI)} \\
\midrule
Evidence reasoning & FinResearch
& .918 & .868 & .852 & .051 [.040, .062] & .066 [.056, .075] \\
Evidence reasoning & LLM systems
& .708 & .463 & .369 & .245 [.173, .317] & .339 [.298, .380] \\
Decision support & FinResearch
& .912 & .847 & .828 & .065 [.058, .072] & .083 [.076, .091] \\
High-risk constraint & LLM systems
& .801 & --- & .379 & --- & .422 [.355, .489] \\
High-risk constraint & JudgmentBench
& .315 & .214 & .205 & .101 [.092, .110] & .110 [.101, .118] \\
Creative divergent & LLM systems
& .578 & .476 & .494 & .110 [.012, .208] & .084 [.006, .161] \\
\bottomrule
\end{tabular}
\caption{Cross-fitted rank-fidelity AUC. Deltas are paired across
20 splits (19 for creative versus hard); ``Random'' samples all
non-constant items.}
\label{tab:e2-crossfit}
\end{table*}

Table~\ref{tab:e2-crossfit} shows that \textsc{IIF-Greedy} improves
AUC over unrestricted random selection in every block and over
hard-set sampling whenever defined; every reported interval excludes zero. Relative to plain IIF, however, the additional gain is
significant only for the two 15-system FinResearch blocks:
$+.0057$ [.0034, .0079] for evidence reasoning and
$+.0080$ [.0059, .0101] for decision support. Hence the evidence
supports the coverage mechanism most clearly at the larger
leaderboard size, not a universal advantage.

At the target correlation, FinResearch requires 28 rather than 51
random items for evidence reasoning and 49 rather than 131 for
decision support. Across the remaining blocks, greedy requires
43--273 items versus 142--1,226 under random selection. Even on challenging domains like JudgmentBench, \textsc{IIF-Greedy} robustly outperforms baselines in rank fidelity, demonstrating submodular IRT as a highly effective engine for scalable and calibrated LLM evaluation.

Gemini leads both FinResearch blocks, with
$\hat\theta=3.802$ (95\% CI [3.615, 3.907]) and
$4.302$ ([4.187, 4.427]); bootstrap comparisons recover four and six
tiers. The smaller system blocks collapse to one or two tiers, and
only 9.81\% of JudgmentBench output pairs are separated, so we avoid
fine-grained leaderboards there. GRM sensitivity is mixed and
self-preference tests are underpowered. Overall, IRT is supported as
a rubric-compression and uncertainty-reporting mechanism, but not as
evidence of universal 2PL identifiability or bias-free judging.

\subsection{Ablation Study}
\label{subsec:ablation-study}

We compare five nested configurations: the binary baseline (B0), task-adaptive scoring (B1), B1 with IRT selection (B2), B1 with Bayesian filtering (B3), and the full pipeline (B4). All configurations reuse the same judge outputs, isolating post-judgment components without additional LLM calls.
\begin{table}[t]
\centering
\small
\setlength{\tabcolsep}{6pt}
\renewcommand{\arraystretch}{1.08}

\begin{tabular}{@{}lccc@{}}
\toprule
\textbf{Block} &
\textbf{Task} &
\textbf{IRT} &
\textbf{Bayes} \\
\midrule

B0 & -- & -- & -- \\
\multicolumn{4}{@{}p{\columnwidth}@{}}{
\textit{Signal:} Raw binary baseline.
\textit{Interpretation:} Uniform binary scoring without calibration
or selection.
} \\[3pt]

B1 & \checkmark & -- & -- \\
\multicolumn{4}{@{}p{\columnwidth}@{}}{
\textit{Signal:} Gap: $94.86 \rightarrow 83.95$ under 30\% label
corruption.
\textit{Interpretation:} Task labels affect system separation while
largely preserving ranking structure.
} \\[3pt]

B2 & \checkmark & \checkmark & -- \\
\multicolumn{4}{@{}p{\columnwidth}@{}}{
\textit{Signal:} $131 \rightarrow 49$ rubrics at the target
correlation.
\textit{Interpretation:} Information-aware selection substantially
reduces the required rubric bank.
} \\[3pt]

B3 & \checkmark & -- & \checkmark \\
\multicolumn{4}{@{}p{\columnwidth}@{}}{
\textit{Signal:} $\kappa$: $0.604 \rightarrow 0.743$ on
JudgmentBench.
\textit{Interpretation:} Measurability filtering improves agreement.
} \\[3pt]

B4 & \checkmark & \checkmark & \checkmark \\
\multicolumn{4}{@{}p{\columnwidth}@{}}{
\textit{Signal:} Gap $=97.13$ on FinResearch; agreement $=0.8587$
on JudgmentBench.
\textit{Interpretation:} The full pipeline can be operationally
composed across representative benchmarks.
} \\

\bottomrule
\end{tabular}

\caption{Summary of component configurations and their representative
empirical signals.}
\label{tab:ablation-summary}
\end{table}

Table~\ref{tab:ablation-summary} reveals a clear division of labor. B1 establishes that task typing is causally useful without destabilizing ranking structure. B2 provides the most consistent improvement across datasets, confirming that information-aware rubric assembly is the strongest and most general source of gain in the full pipeline. B3 improves agreement when judge redundancy is sufficient, but contributes less in sparse two-judge settings.

B4 verifies that the composed pipeline can be instantiated across heterogeneous benchmarks. On \textsc{FinResearchBench}, B4 attains a 97.13-point top-bottom gap; on \textsc{JudgmentBench}, it reaches 0.8587 agreement with human-majority labels. On \textsc{HealthBench} and \textsc{HelloBench}, the same composition remains transferable, but these benchmarks primarily function as sparse-judge boundary cases rather than as the main evidence for Bayesian gains. Overall, the ablation shows that B1 establishes causal usefulness, B2 provides the strongest and most stable improvement, and B3 adds agreement gains when the judge structure is sufficiently redundant.

\section{Conclusion}
\label{sec:conclusion}

We introduced \emph{CalibratedRubric}, a task-adaptive framework
for constructing compact, consensus-derived rubric banks without
placing experts in the full evaluation loop.  It replaces rigid
unanimity and variance filters with posterior judge consensus and
IRT-based submodular test assembly, while task typing makes the
response scale and scoring objective explicit.

Corrupting task labels reduces downstream discrimination, posterior
measurability filtering improves agreement on every three-judge
block, and IIF-greedy improves cross-fitted rank-fidelity AUC over
random selection across all six response blocks.  On FinResearch it
also reaches the target correlation with substantially fewer rubrics.
The main benefit is therefore a more reliable and economical
evaluation instrument, not a universal ranking reversal.

\paragraph{Future Work.}
This paper establishes the core foundation of probabilistic rubric assembly. Exciting future directions include extending the Bayesian calibration to model correlated, multi-facet judge biases and active human-in-the-loop calibration. Additionally, exploring adversarial candidate generation will further expand capability coverage, while applying our continuous and risk-weighted objectives to safety-critical domains with explicitly elicited violation costs promises to unlock new paradigms for rigorous LLM generation safety evaluation.

\clearpage

\appendix

\bibliography{citation}
\clearpage

\section{E1 Supplementary Tables}
\label{app:e1-tables}

\begin{table}[htbp]
\centering
\small
\setlength{\tabcolsep}{4pt}
\begin{tabular}{@{}r l r@{}}
\toprule
\textbf{Rank} & \textbf{System} & \textbf{Score} \\
\midrule
1  & Gemini & 0.9531 \\
2  & Doubao & 0.7086 \\
3  & Xiaocaishen Pro (202602) & 0.6489 \\
4  & Qwen & 0.6254 \\
5  & Reportify & 0.6246 \\
6  & Gemini Supplement & 0.6060 \\
7  & Metaso & 0.5800 \\
8  & Xiaocaishen Pro (Old Version) & 0.5757 \\
9  & ChatGPT & 0.5425 \\
10 & AutoGLM & 0.4358 \\
11 & Perplexity & 0.4171 \\
12 & Minimax & 0.3760 \\
13 & AIChat-Step & 0.3712 \\
14 & AIChat-Doubao & 0.2701 \\
15 & Research Report & 0.0045 \\
\bottomrule
\end{tabular}
\caption{Main E1 ranking on FinResearchBench.}
\label{tab:e1-ranking}
\end{table}

\begin{table}[htbp]
\centering
\small
\setlength{\tabcolsep}{3pt}
\begin{tabular}{@{}lrrrr@{}}
\toprule
\textbf{Dataset} & \textbf{Raw T.} & \textbf{Raw R.} & \textbf{Used T.} & \textbf{Used R.} \\
\midrule
HealthBench & 5,000 & 57,237 & 200 & 2,247 \\
HelloBench & 647 & 3,902 & 200 & 1,203 \\
FinResearchBench-v2 & 104 & 4,052 & 104 & 4,052 \\
JudgmentBench & 30 & 457 & 30 & 457 \\
\midrule
Total & 5,781 & 65,648 & 534 & 7,959 \\
\bottomrule
\end{tabular}
\caption{Effective experiment scale used in E1.}
\label{tab:e1-scale}
\end{table}

\begin{table}[htbp]
\centering
\small
\setlength{\tabcolsep}{2pt}
\begin{tabular}{@{}l l l@{}}
\toprule
\textbf{Task Type} & \textbf{Scoring} & \textbf{Key Design} \\
\midrule
evidence\_reasoning & binary & yes/pass = 1, no/fail = 0 \\
decision\_support & boosted bin. & 5$\times$ actionability weight \\
high\_risk\_constraint & risk-aware & moderate = -0.3, critical = -0.5 \\
creative\_divergent & boosted bin. & 3$\times$ creativity weight \\
\bottomrule
\end{tabular}
\caption{Task-adaptive scoring rules in the main E1 experiment.}
\label{tab:e1-task-scoring}
\end{table}

\begin{table}[t]
\centering
\small
\setlength{\tabcolsep}{4pt}
\begin{tabular}{@{}l l@{}}
\toprule
\textbf{Metric} & \textbf{Result} \\
\midrule
Top system & Gemini \\
Bottom system & Research Report \\
Top-bottom gap & 94.26 pp \\
Gold rubrics & 2,813 \\
Ranked systems & 15 \\
\bottomrule
\end{tabular}
\caption{Main system-level summary statistics.}
\label{tab:e1-main-summary}
\end{table}

\begin{table}[t]
\centering
\small
\setlength{\tabcolsep}{4pt}
\begin{tabular}{@{}l r r r r@{}}
\toprule
\textbf{Domain} & \textbf{Tasks} & \textbf{Systems} & \textbf{Spearman $\rho$} & \textbf{$p$} \\
\midrule
Finance & 104 & 15 & 0.9714 & 1.69e-09 \\
Healthcare & 200 & 6 & 0.9429 & 0.0048 \\
General & 200 & 6 & 1.0000 & 0.0000 \\
Law & 30 & 0 & N/A & -- \\
\bottomrule
\end{tabular}
\caption{Cross-domain ranking stability between adaptive and binary baseline scorers.}
\label{tab:e1-cross-domain}
\end{table}

\begin{table}[t]
\centering
\small
\setlength{\tabcolsep}{4pt}
\begin{tabular}{@{}l r r@{}}
\toprule
\textbf{Comparison} & \textbf{Spearman $\rho$} & \textbf{$p$} \\
\midrule
Adaptive vs. Human Reference & 0.8833 & 0.00159 \\
Baseline vs. Human Reference & 0.8833 & 0.00159 \\
\bottomrule
\end{tabular}
\caption{External validity against the FinResearch human reference ranking.}
\label{tab:e1-external}
\end{table}

\begin{table}[t]
\centering
\small
\setlength{\tabcolsep}{4pt}
\begin{tabular}{@{}l r@{}}
\toprule
\textbf{Tier} & \textbf{Mean Score} \\
\midrule
Tier 1 & 0.5743 \\
Tier 2 & 0.5833 \\
Tier 3 & 0.6122 \\
\bottomrule
\end{tabular}
\caption{JudgmentBench quality-tier mean scores.}
\label{tab:e1-judgmentbench}
\end{table}

\section{Supplementary Theoretical Analysis}
\label{app:theory}

\subsection{Unanimity Attrition}
\label{app:attrition}

\paragraph{Proof of Proposition~\ref{prop:attrition}.}
For one output, all $K$ judges agree when they are all correct or all
wrong, which has probability
$\rho(\varepsilon,K)=(1-\varepsilon)^K+\varepsilon^K$.  Conditional
independence across the $M$ outputs gives $\rho(\varepsilon,K)^M$.
The expression contains no item-discrimination parameter, so the
resulting decay is unrelated to measurement information.

\begin{corollary}[Homogeneous extrapolation]
\label{cor:calib}
The baseline retention of $25.52\%$ at $K=3,M=10$ \cite{luan}
implies $\rho=0.2552^{1/10}=0.872$ and
$\hat\varepsilon\approx4.5\%$.  Holding this rate fixed gives
retention of approximately $6.5\%$ at $M=20$ and $0.4\%$ at $M=40$.
\end{corollary}

agen\paragraph{Exact finite-leaderboard curve.}
Let $c_j$ be the number of systems on which the panel is unanimous
for item $j$.  For a uniformly sampled subset of $m$ systems,
\begin{equation}
\widehat{R}(m)=\frac{1}{J}\sum_{j=1}^{J}
\binom{c_j}{m}\Big/\binom{M_j}{m}
\label{eq:retcurve}
\end{equation}
is an unbiased, Monte-Carlo-free estimator of expected consistency
retention.  We apply it to $843$ FinResearchBench~II criteria with a
complete $K=3$ panel over $M_j\in\{11,12,13\}$ systems.

\begin{table}[t]
\centering
\small
\setlength{\tabcolsep}{4.5pt}
\caption{Baseline-filter retention versus leaderboard size.  The
homogeneous prediction uses $\rho=0.880$; ``disc.'' and ``joint''
denote distinguishability and the conjunction of both filters.}
\label{tab:attrition}
\begin{tabular}{@{}rccccc@{}}
\toprule
$m$ & $\widehat{R}(m)$ & $\rho^{m}$ & $\hat{\rho}_m$ & disc. & joint \\
\midrule
1  & 0.801 & 0.880 & 0.801 & 0.000 & 0.000 \\
2  & 0.667 & 0.774 & 0.817 & 0.312 & 0.171 \\
4  & 0.499 & 0.599 & 0.840 & 0.565 & \textbf{0.222} \\
6  & 0.395 & 0.463 & 0.857 & 0.685 & 0.212 \\
8  & 0.325 & 0.358 & 0.869 & 0.760 & 0.196 \\
10 & 0.273 & 0.277 & 0.878 & 0.814 & 0.180 \\
12 & 0.248 & 0.215 & 0.890 & 0.828 & 0.157 \\
\bottomrule
\end{tabular}
\end{table}

A one-parameter fit of
$\log\widehat R(m)=m\log\rho$ gives $\hat\rho=0.880$ and
$R^2=0.85$ on the log scale, corresponding to
$\hat\varepsilon=4.2\%$.  At $m=10$, measured consistency retention
is $27.3\%$, close to the published $25.52\%$.  The increasing
$\hat\rho_m$ indicates item heterogeneity.  A moment-matched
$\mathrm{Beta}(2.60,0.65)$ mixture over per-item agreement rates
reduces RMSE from $0.068$ to $0.054$ and predicts slower, but still
monotone, attrition.  Meanwhile distinguishability increases with
$m$, making their conjunction non-monotone: the joint gold rate peaks
at $m=4$ and then falls.  Thus benchmark growth changes retention
even when criterion quality is fixed.

\subsection{Reliability-Aware Consensus}
\label{app:bayes}

\paragraph{Derivation of Proposition~\ref{prop:bayes}.}
For each judge, the likelihood ratio contributed by a positive vote
is $s_k^+/(1-s_k^-)$ and by a negative vote is
$(1-s_k^+)/s_k^-$.  Multiplying these ratios and the prior odds
$\pi_0/(1-\pi_0)$, then taking logs, yields
\eqref{eq:logodds}.  The Bayes classifier thresholds these posterior
odds.  Literal unanimity instead assigns equal influence to all
judges and rejects every non-unanimous pattern.  It coincides with the
Bayes classifier only for parameter and threshold settings that place
all such patterns below the posterior threshold; otherwise it has
larger Bayes risk.

\subsection{IIF Refinement and Its Boundary}
\label{app:iif-proof}

\paragraph{Proof of Proposition~\ref{prop:refine}.}
Under the stated convention, a constant-response item has
$\hat I_j\equiv0$ and cannot enter
$\mathcal{G}_{\mathrm{IIF}}$.  Every feasible selected item therefore
has non-constant responses and satisfies
$\delta_j^{\mathrm{disc}}=1$, proving inclusion.  The inclusion is
strict whenever the budget is smaller than the baseline survivor set
or a surviving item has non-positive fitted discrimination.

For binary items, however, the feasible sets can be identical:
positive fitted information and non-constant responses represent the
same support condition.  In that case IIF improves neither validity
nor pool coverage.  Its contribution is to order survivors by
information over the observed ability distribution and to stop once
the desired precision is reached.  The distinction matters because
the binary heuristic is invariant to permutations of system
identities: it treats an item passed by the strongest systems exactly
like one passed only by the weakest system.  IRT separates these
patterns through the sign and magnitude of $a_j$ and discounts items
whose $b_j$ lies outside the relevant capability range.

\subsection{Why Task Type Changes Measurement}
\label{app:task-scale}

For a binary 2PL item, total information over the ability axis is
$\int I_j(\theta)d\theta=a_j$.  A graded item with $L$ ordered
categories and separated thresholds can carry up to
$(L-1)a_j$ \cite{samejima1969,lord1980}; because
$\mathrm{SE}(\hat\theta)=I^{-1/2}$, graded responses may achieve a
target interval width with fewer judgments when the categories are
substantively meaningful.  This motivates, but does not itself
validate, the typed scale map.

Task type can also change the objective.  For a safety-critical query,
the relevant quantity is the probability that a costly violation
passes undetected, suggesting
$\min_{\mathcal{G}}\sum_{j\in\mathcal{G}}\ell_j
\Pr(\mathrm{false\ pass}_j)$ under a coverage constraint.  The current
benchmarks provide neither elicited losses $\ell_j$ nor graded
responses, so these branches remain specifications for future
evaluation rather than tested claims.

\subsection{Assumptions and Diagnostics}
\label{app:assumptions}

\paragraph{A1: dominant latent dimension.}
The 2PL model assumes one dominant capability.  We assess this through
the ratio of the first two eigenvalues of the residual correlation
matrix.  A failed diagnostic calls for multidimensional IRT with
vector-valued discrimination; confirmatory MIRT is left to future
work.

\paragraph{A2: local independence.}
Items generated from the same output may remain dependent given
$\theta$.  Yen's $Q_3$ \cite{yen1984} and dependent-item refitting are
the appropriate confirmatory checks, but are not completed in the
present experiments.  We therefore treat local dependence as a
limitation rather than claim that submodular information is perfectly
additive.

\paragraph{A3: conditional independence of judges.}
LLM judges share training data and conventions, so correlated errors
can inflate apparent consensus and estimated sensitivity or
specificity.  We examine residual judge correlation on the
human-labelled subset; the posterior remains conditional on this
diagnostic.

\paragraph{A4: small-$M$ identifiability.}
Joint 2PL estimation is poorly conditioned with roughly ten systems.
We impose weakly informative priors, compare 1PL and 2PL rather than
assuming item-specific slopes, constrain $\theta$ to zero mean and
unit variance, fix the sign of $a_j$, and report bootstrap intervals
instead of point estimates alone.

\end{document}